\documentclass[letterpaper, 10 pt, conference]{ieeeconf}  % Comment this line out if you need a4paper

\IEEEoverridecommandlockouts                              % This command is only needed if 
\usepackage{cite}
\usepackage{amsmath,amssymb,amsfonts}
\usepackage{algorithmic}
\usepackage{graphicx}
\usepackage{textcomp}
\usepackage{xcolor}
\usepackage{etoolbox}
\usepackage{graphicx}
\usepackage[skip=0.333\baselineskip]{caption}
\usepackage{subcaption}
\usepackage{multirow}
\usepackage{multicol}
\usepackage{hyperref}

\title{\LARGE \bf
HawkDrive: A Transformer-driven Visual Perception System for Autonomous Driving in Night Scene
}

\author{Ziang Guo$^{1}$, Stepan Perminov$^{2}$, Mikhail Konenkov$^{3}$ and Dzmitry Tsetserukou$^{4}$% <-this % stops a space
\thanks{$^{1}$Ziang Guo is with the Intelligent Space Robotics Laboratory, Center for Digital Engineering, Skolkovo Institute of Science and Technology, Moscow, Russia
        {\tt\small ziang.guo@skoltech.ru}}%
\thanks{$^{2}$Stepan Perminov is with the Intelligent Space Robotics Laboratory, Center for Digital Engineering, Skolkovo Institute of Science and Technology, Moscow, Russia, and with the LLC IntegraNT, Moscow, Russia
        {\tt\small stepan.perminov@skoltech.ru}}%
\thanks{$^{3}$Mikhail Konenkov is with the Intelligent Space Robotics Laboratory, Center for Digital Engineering, Skolkovo Institute of Science and Technology, Moscow, Russia
        {\tt\small mikhail.konenkov@skoltech.ru}}%
\thanks{$^{4}$Dzmitry Tsetserukou is with the Intelligent Space Robotics Laboratory, Center for Digital Engineering, Skolkovo Institute of Science and Technology, Moscow, Russia
        {\tt\small d.tsetserukou@skoltech.ru}}%
}

\begin{document}

\maketitle
\thispagestyle{empty}
\pagestyle{empty}

%%%%%%%%%%%%%%%%%%%%%%%%%%%%%%%%%%%%%%%%%%%%%%%%%%%%%%%%%%%%%%%%%%%%%%%%%%%%%%%%
\begin{abstract}

Many established vision perception systems for autonomous driving scenarios ignore the influence of light conditions, one of the key elements for driving safety. To address this problem, we present HawkDrive, a novel perception system with hardware and software solutions. Hardware that utilizes stereo vision perception, which has been demonstrated to be a more reliable way of estimating depth information than monocular vision, is partnered with the edge computing device Nvidia Jetson Xavier AGX. Our software for low light enhancement, depth estimation, and semantic segmentation tasks, is a transformer-based neural network. Our software stack, which enables fast inference and noise reduction, is packaged into system modules in Robot Operating System 2 (ROS2). Our experimental results have shown that the proposed end-to-end system is effective in improving the depth estimation and semantic segmentation performance. Our dataset and codes will be released at \href{https://github.com/ZionGo6/HawkDrive}{\textit{https://github.com/ZionGo6/HawkDrive}}.

\end{abstract}

\section{Introduction}

\subsection{Motivation}

Visual perception of self-driving vehicles in low light conditions remains a challenge. Insufficient and patchy illumination in the night driving environment can cause considerable interference to cameras, thus adversely affecting the performance of neural network \cite{sharma2021nighttime}. \par Besides, data-driven monocular depth estimation for self-driving is susceptible to adversarial conditions such as image noise, occlusions, variations in lighting, and surface reflections, while the stereo camera is demonstrated with more  sufficient accuracy compared to the monocular one \cite{smolyanskiy2018importance}, \cite{xu2023unifying}.
Based on the data processing from stereo camera, a flexible edge computing method should be developed since it performs capturing and processing data as close to its source or end user as possible to empower the perception system with low latency for autonomous driving. For more reliable downstream task performance, as the learning-based perception models are developed less vigorously, it is also necessary to deploy state-of-the-art neural networks on the edge computation devices for inference and evaluation.

\begin{figure}[t]
    \centering
    \begin{subfigure}{.47\linewidth}
        \includegraphics[width=\linewidth]{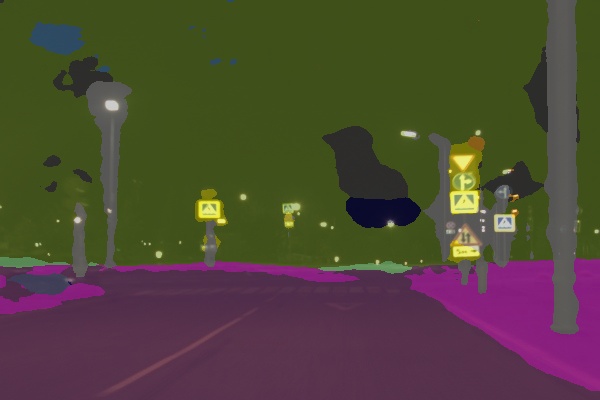}
        \subcaption{Semantic segmentation of night image.}
    \end{subfigure}
    \begin{subfigure}{.47\linewidth}
        \includegraphics[width=\linewidth]{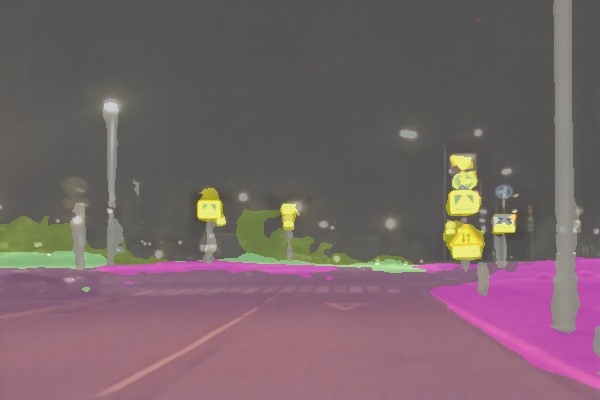}
        \subcaption{Semantic segmentation of enhanced image.}
    \end{subfigure}

    \caption{Visual comparison of semantic segmentation performance from SegFormer module.}
    \label{fig:seg_visual_comparison}
\vspace{-2em}
\end{figure}

\subsection{Problem Statement}

Errors in depth estimation and image noise caused by unfavorable sensing conditions are critical factors of failure in perception, navigation, and planning \cite{liu2023improving}. Passive sensors, such as RGB cameras, can acquire noisy, low-dynamic range and blurry images. Therefore, a versatile pipeline for post-processing raw sensor data is required. Moreover, a satisfactory processing speed is needed in order to adapt to various driving situations. In particular, it is challenging for cameras to discover full details in low-light conditions. A few prior works have been proposed to enhance night-time camera data, such as TodayGAN \cite{anoosheh2019night}, EnlightenGAN \cite{jiang2021enlightengan}. Nevertheless, these techniques are not able to manage high dynamic range scenes, as they are unable to maintain enough details for driving scenes that have a variety of dynamic ranges of sight and illumination conditions due to the non-uniform boosting areas of the entire frame \cite{zheng2020forkgan}.

\subsection{Related Work}

Many modern binocular or monocular visual perception systems conceive ideal raw image input, but such consideration does not provide thorough robustness for downstream missions. Faisal et al. \cite{faisal2022depth} developed an algorithm for depth estimation of moving objects based on feature detection, extraction and matching. Königshof et al.\cite{konigshof2019realtime} proposed a car-oriented 3D detection and pose estimation method via semantic information, disparity and geometry constraints to recover the 3D bounding boxes. Zaarane et al. \cite{zaarane2019vehicle} designed a vehicle-to-vehicle measurement module for preventing traffic accidents. Kemsaram et al. \cite{kemsaram2022model} devised a stereo vision system for obstacle motion estimation among cooperative autonomous vehicles based on disparity information. \par Each method did not consider the light influence on their systems where the disparity map is obtained as the key processing of stereo image input. However, illumination conditions challenge the sensors' reliability, as well as affect the disparity estimation \cite{almalioglu2022deep}. \par Low light enhancement task has achieved adaptable development. Yeonjun Bang et al. \cite{bang2022EV} proposed their work of reliable image-to-image translation but focused on electric vehicle (EV) inlet detection case. Whilst in the work of Savinykh et al.\cite{savinykh2022darkslam}, towards indoor environments, a generative adversarial network (GAN) was implemented to enhance different illumination levels of images. However, the detected features can be influenced by the noise of the processed images. 

\subsection{Contribution}

In this paper, we present, HawkDrive, a stereo camera system developed for acquiring depth and semantic maps for autonomous driving in night scene with a low light enhancement framework. Images are captured from different perspectives from a stereo pair of cameras, which can provide information about the distance of objects from the cameras \cite{moll2008truthing}. To allow for quick parameter tuning, different camera configurations are created with adjustable properties from the camera kernel drivers via ROS2 node configurations \cite{ROS2_Composition}. \par The image data obtained from the stereo cameras requires accurate hardware synchronization and low latency. Thus, hardware trigger cables and Nvidia Jetson Xavier AGX are partnered within an adjustable in-car structure for experimental validation combining 3D-printed computation device base, camera holders and pneumatic support. \par For our software, Signal-to-Noise-Ratio-aware (SNR-aware) transformers and convolutional models \cite{xu2022snr}, are adapted to perform low light enhancement for the nighttime driving scene with prior knowledge input powered by SegFormer-based semantic segmentation network \cite{xie2021segformer} allowing the enhancement based on both physical and semantic information. To evaluate the enhancement effect, depth estimation modules based on Vision Transformers for Dense Prediction (DPT) \cite{ranftl2021vision} and Unimatch\cite{xu2023unifying} are adapted on this stereo camera - Nvidia Jetson Xavier AGX system.
 
% \begin{figure}
% \centering
% \includegraphics[width=0.48\textwidth,height=\textheight, keepaspectratio]{figs/stereo_camera_structure.png}
% \caption{Decomposition of the HawkDrive structure.}
% \label{fig:stereostructure}
% \vspace{-1.5em}
% \end{figure}

% \begin{figure}
% \centering
% \includegraphics[width=0.3\textwidth,height=\textheight, keepaspectratio]{figs/installation.png}
% \caption{HawkDrive installed in the self-driving car.}
% \label{fig:stereo_install}
% \vspace{-1.5em}
% \end{figure}

\section{System Overview}

\begin{figure*}[t]
% \vspace{-1em}
\centering
\includegraphics[width=0.7\textwidth,height=\textheight, keepaspectratio]{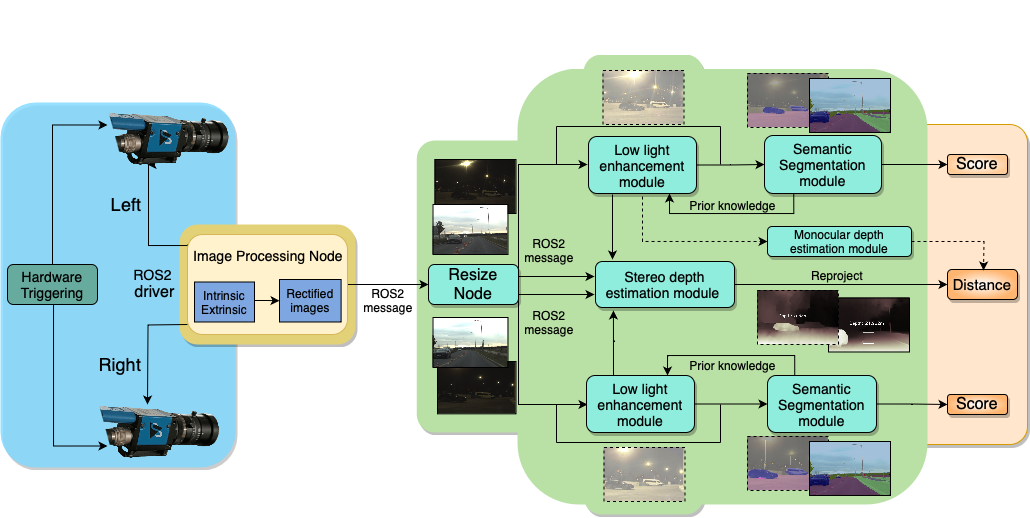}
\caption{System overview. Here is the workflow for night driving scene, ROS2 messages of resized image are published to low light enhancement and segmentation modules to perform semantic knowledge based enhancement. Then the enhanced images are conveyed to other modules for corresponding tasks.}
\label{fig:system}
\vspace{-1.5em}
\end{figure*}

\subsection{Hardware}

Two global-shutter The Imaging Source DFK 33UX250 cameras with 2.3 mm machine vision lens Computar T2314 FICS, CMOS Pregius Sony IMX250 sensor are used as capture modules which can provide a ($2,448\times2,048$) 5MP resolution and 12 bit dynamic range. Two global shutter cameras with a 67.12 cm baseline capture images of the entire sensor at once, ensuring that all pixels are exposed simultaneously \cite{albl2020two}. Utilizing a synchronized pair of global shutter cameras eliminates the rolling shutter effect and ensures accurate timing of events. \par Nvidia's Jetson Xavier AGX is an embedded computer that has been created for AI applications. We use one of these devices as the computation module of our system. The Jetson Xavier AGX supports accelerated inference of deep learning algorithms, making it suitable for dynamic scene applications such as autonomous cars. \par Therefore, the combination of the Nvidia Jetson Xavier AGX and stereo vision is able to offer a promising solution for autonomous driving scenes \cite{jetsonforDL}.

\subsection{Driver Build}

ROS2 is chosen as the platform for communication and camera data processing \cite{ROS2}. This central platform initiates the trigger mode from TheImagingSource DFK 33UX250 cameras, hardware and software level synchronized stereo images are both able to be obtained. With the parameter adjustments from the node configuration file such as shutter speed and gain values, the adjusted images can be recorded for sufficient information extraction. \par Passive sensors, such as cameras, rely on analyzing 3D visual information captured by the sensors synchronously and working with algorithms to estimate distances. Therefore, for synchronized image capture, hardware triggering with a current amplifier is used between the Nvidia Jetson Xavier AGX I/O pins and the camera trigger inputs. The built driver is applied to capture configured images for corresponding tasks. 

\begin{figure}
\vspace{1em}
\centering
\includegraphics[width=0.45\textwidth]{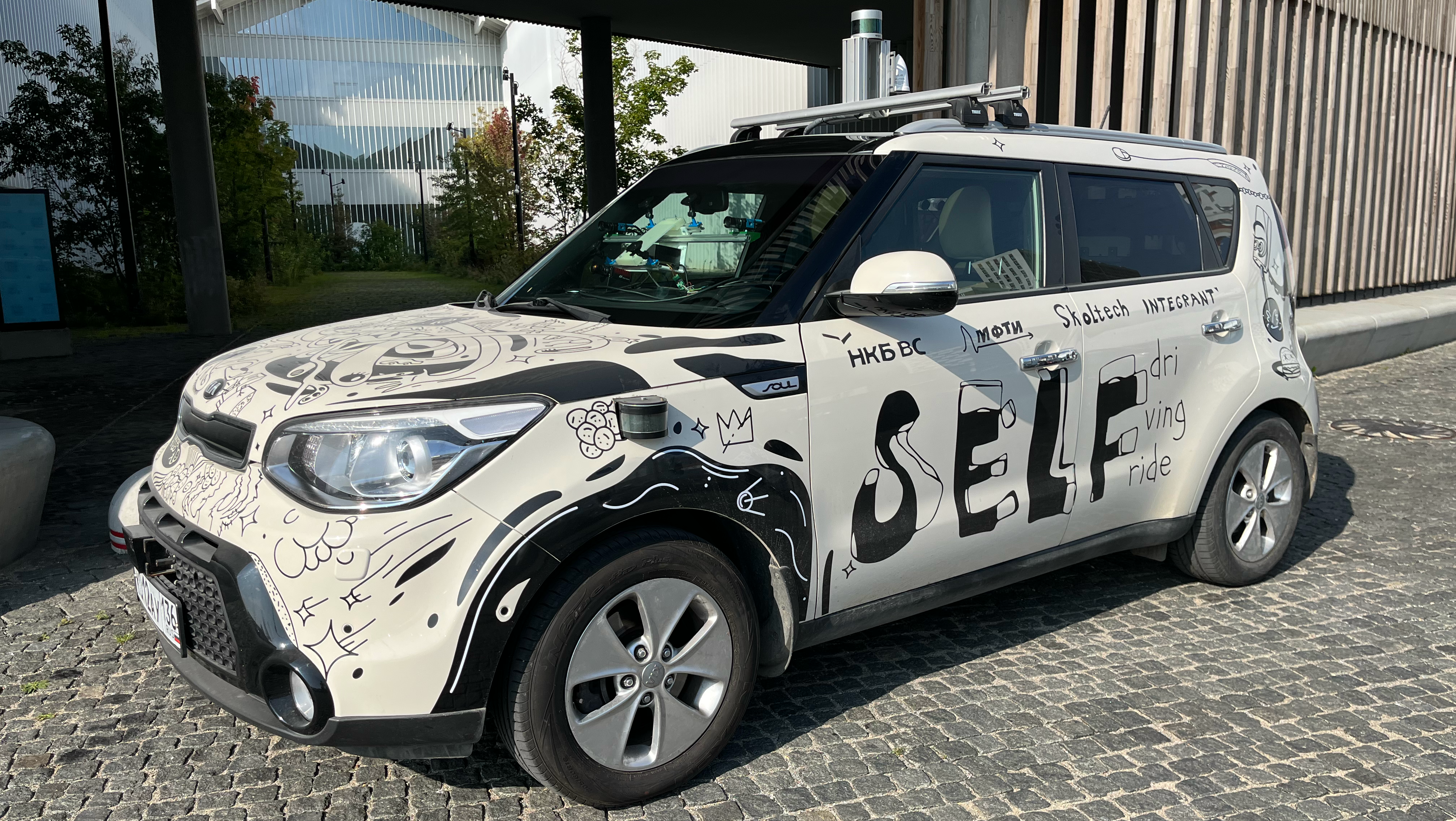}
\caption{Experimental Self-Driving Car.}
\label{fig:sdc}
\vspace{-1.5em}
\end{figure}

\subsection{Pipeline}

Initiating the camera ROS2 driver, the stereo images from the left and right cameras are saved as dataset for stereo calibration preparation. Then, using a stereo calibration algorithm, the stereo calibration results are obtained. For processing the calibration matrix, two methods are developed. First, the calibration matrix is integrated into the ROS2 node configuration which will be done by filling the Camera\_Info messages. Second, the calibration results are integrated into the ROS2 package with OpenCV \cite{opencv}. \par Receiving the current from Jetson pins through an amplifier, the two cameras are connected via trigger cables. The pulse output is controlled by the trigger module to control the camera frame rate and exposure time input. Then a ROS2 package for receiving hardware synchronized images from left and right camera is built, transforming the image data to ROS2 messages and publishing the synchronized messages via ROS2 nodes. For downstream processing, the message subscribers and synchronizers are respectively established. Through the built driver, rectified images from the stereo camera are output, while they are then subscribed by the low light enhancement module, depth prediction and semantic segmentation module. All the modules work in customized docker containers \cite{docker}, \cite{isaacROS} and the whole system overview can be concluded as Fig. \ref{fig:system}.

\section{Methodology}

\subsection{Signal-to-Noise-Ratio-aware Low Light Enhancement}

Recent research has shown that low light images are usually not suitable for human perception \cite{xu2022snr}. Similarly, when low light images are directly used as the input to a perception system, the downstream vision tasks could be affected \cite{xia2023image}. \par Starting from better overall image enhancement via adaptive considerations of different regions in low light images, Xu et al. studied the relation between signal and noise in image space by exploring Signal-to-Noise-Ratio (SNR). \par Image noise is treated as a discontinuous transition between adjacent pixels in the spatial domain. The SNR map 
%$S \in {\rm I\!R}^{H \times W}$% 
obtained from the distance between the night image and an associated gray-scale image is used to guide the fusion of long-range (transformer processing results) and short-range (convolutional neural network processing results) features (i.e. the fusion of local information and global information). \par
\begin{figure}
    \centering
    \begin{subfigure}{.47\linewidth}
        \centering
        \includegraphics[width=\linewidth]{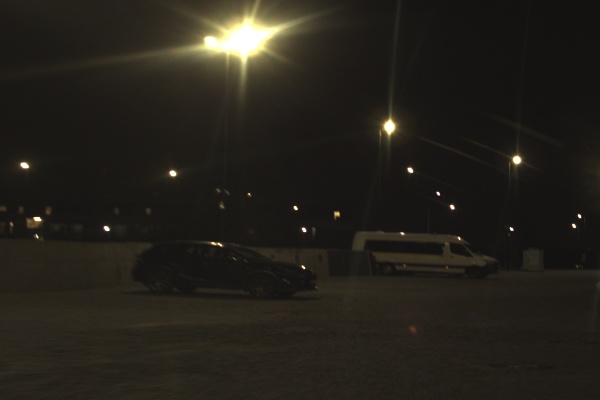}
         \caption{Input dark camera image.}
    \end{subfigure}
    \begin{subfigure}{.47\linewidth}
        \centering
        \includegraphics[width=\linewidth]{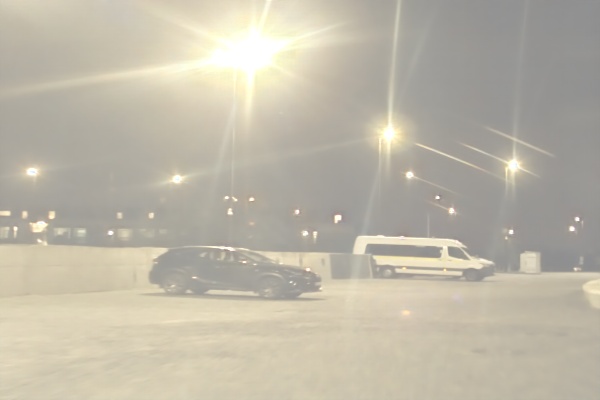}
        \caption{Output enhanced image.}
    \end{subfigure}
    
    \caption{SNR-aware low light enhancement module processing.}
    \label{fig:snr_example}
\vspace{-0.5em}
\end{figure}

 Our resize node resizes the raw image to a resolution of $600\times400$, which is then sent to the SNR-aware enhancement module to produce the enhanced image. In Fig. \ref{fig:snr_example}, our dataset's results show a considerable improvement in the night scene, while the details of vehicles, pedestrians, and roads were preserved.

\begin{figure}
    \centering
    \begin{subfigure}{.3\linewidth}
        \centering
        \includegraphics[width=\linewidth]{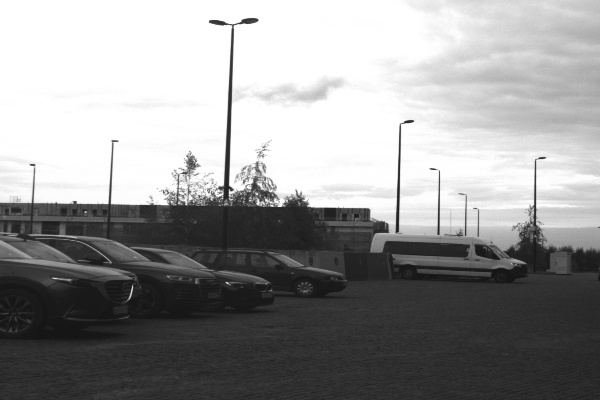}
        \caption{Input left  \\ camera image.}
    \end{subfigure}
    \begin{subfigure}{.3\linewidth}
        \centering
        \includegraphics[width=\linewidth]{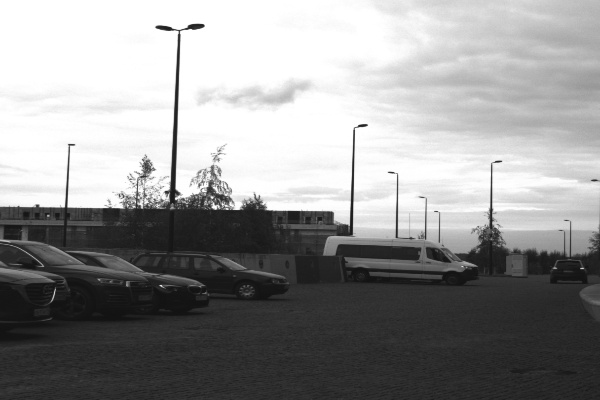}
        \caption{Input right \\ camera image.}
    \end{subfigure}
    \begin{subfigure}{.3\linewidth}
        \centering
        \includegraphics[width=\linewidth]{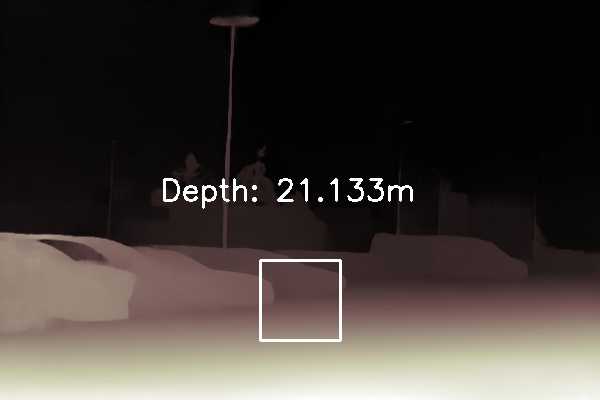}
        \caption{Output depth \\ image.}
    \end{subfigure}

    \caption{Unimatch depth estimation module processing.}
    \label{fig:unimatch_depth_example}
\vspace{-1.5em}
\end{figure}

\begin{figure}[h]
% \vspace{-1em}
\centering
\includegraphics[width=0.45\textwidth]{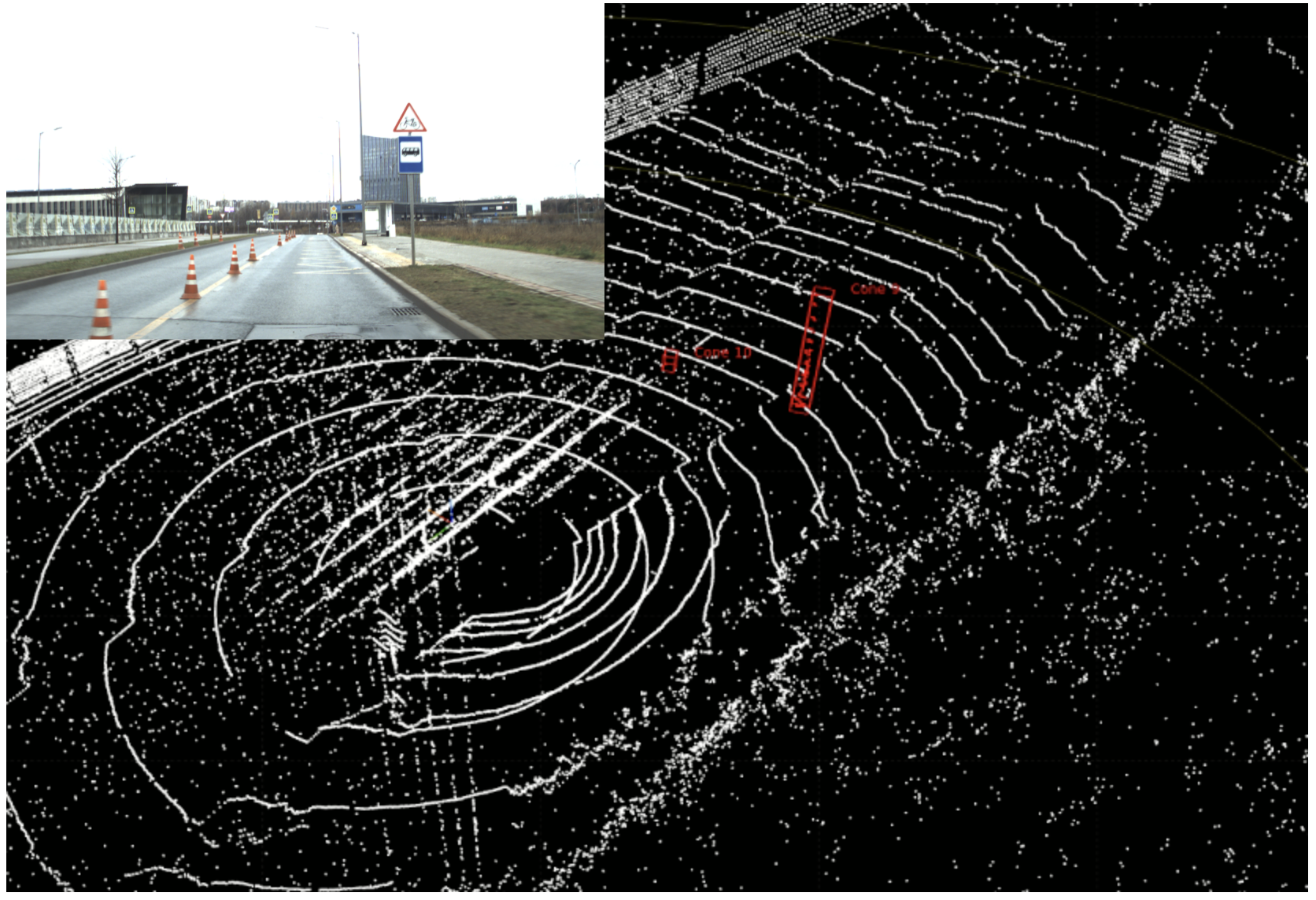}
\caption{Depth estimation obtained from the annotated LiDAR point clouds.}
\label{fig:LiDAR}
\vspace{-2em}
\end{figure}

\vspace{0em}
\subsection{Global Matching Depth Estimation}

Utilizing an approach similar to the classic plane-sweep stereo method \cite{1996plane-sweep}, GMDepth among Unimatch models is assigned for depth estimation. Unimatch \cite{xu2023unifying} is considered as an extended work of GMFlow \cite{xu2022gmflow}, integrating optical flow, stereo matching and depth estimation tasks into one model because stereo matching is their common guideline. The integrated model naturally supports cross-task migration, because the learnable parameters of all tasks are exactly shared among the feature extraction blocks. More specifically, the pre-trained optical flow model can be directly applied to three-dimensional matching and depth estimation tasks.  \par For higher real-time performance, the Unimatch module subscribes to the resized $600\times400$ image messages. Setting the module output as stereo matching, the results from our dataset are shown in Fig. \ref{fig:unimatch_depth_example}. This figure shows the filtered depth at each pixel, by averaging the prediction over a small patch centered at this pixel. \par Ranftl et al. \cite{ranftl2021vision} proposed a dense prediction transformer (DPT) specifically for dense prediction tasks, while the monocular depth estimation can be treated as a dense regression problem. It showed a performance improvement compared to the fully convolutional neural network, so the monocular depth estimation of DPT module is adapted for the cases when one of the stereo cameras has malfunction.

% \begin{figure}[t]
% \vspace{1em}
%     \centering
%     \begin{subfigure}{.47\linewidth}
%         \centering
%         \includegraphics[width=\linewidth]{figs/raw_356.jpg}
%         \caption{Input camera image.}
%     \end{subfigure}
%     \begin{subfigure}{.47\linewidth}
%         \centering
%         \includegraphics[width=\linewidth]{figs/dpt_356.jpg}
%         \caption{Output depth image.}
%     \end{subfigure}

%     \caption{DPT depth estimation module processing.}
%     \label{fig:dpt_depth_example}
% \vspace{-1.5em}
% \end{figure}

\subsection{SegFormer}

SegFormer is used as the semantic segmentation module of our system, due to the limited computational resources of edge devices. Segformer avoids interpolating positional encoding for different inference image resolutions, thus improving the efficiency, accuracy and robustness of the segmentation performance \cite{xie2021segformer}. Besides, SegFormer showed superior performance in handling common corruptions and perturbations such as noise, motion blur and weather influence. \par
\begin{figure*}[t]
\centering
\includegraphics[width=0.7\textwidth,height=\textheight, keepaspectratio]{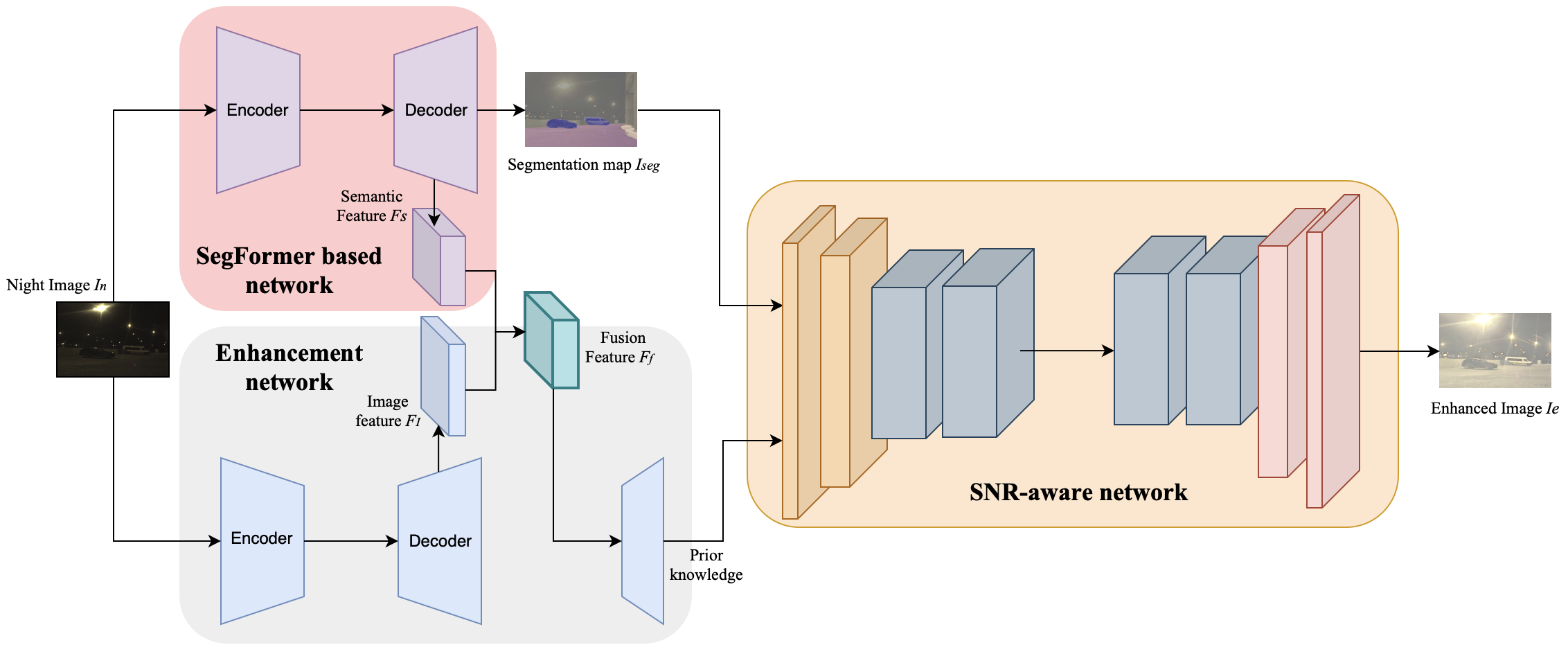}
\caption{Low light enhancement with semantic prior knowledge pipeline. Night image $I_n$ is segmented through SegFormer module to obtain segmentation map $I_{seg}$ and extracted semantic feature $F_s$ while an encoder-decoder enhancement network provides extracted image feature $F_I$. Then the fusion feature $F_f$ is obtained from element-wise operations. Segmentation map $I_{seg}$ and fusion feature $F_f$ are both input towards Signal-to-Noise guided network to get enhanced image $I_e$.}
\label{fig:network}
\vspace{-1.5em}
\end{figure*}

 Meanwhile, the SegFormer module subscribes to raw image input and publishes segmentation map to low light enhancement module as prior knowledge input to refine the boosting results. In Fig. \ref{fig:network}, night image $I_n$ is conveyed to segmentation network and enhancement network to obtain segmentation map $I_{seg}$ and fusion feature $F_f$. The attention map $A$ is computed as follows,

\begin{equation*}
  A = Softmax(W_q(F_s) \times W_k(F_I) / \sqrt{C})    \tag{1}
\end{equation*}
\vspace{-1.5em}
\newline where $W_q$ and $W_k$ mean the weights of query and key. $C$ is the channel of features.
\par The fusion feature $F_f$ is computed as follows,

\begin{equation*}
  F_f = FFN(W_v(F_I) \times A + F_I)    \tag{2}
\end{equation*}
\vspace{-1.5em}
\newline where $FFN$ is the feed-forward networks and $W_v$ is the weight of value.

% \begin{figure}
% \centering
%     \begin{subfigure}{.47\linewidth}
%         \centering
%         \includegraphics[width=\linewidth]{figs/resize_right_373.jpg}
%         \caption{Input camera image.}
%     \end{subfigure}
%     \begin{subfigure}{.47\linewidth}
%         \centering
%         \includegraphics[width=\linewidth]{figs/day_8.jpg}
%         \caption{Output segmented image.}
%     \end{subfigure}
%     \caption{SegFormer module processing.}
%     \label{fig:segformer_example}
% \vspace{-1.5em}
% \end{figure}

\section{Experimental Results}

\subsection{Experimental Setup}

With onboard servers and power supply of the experimental car, shown in Fig. \ref{fig:sdc}, the test dataset was acquired along the same driving routine from daytime and nighttime driving scenes, while two cameras were hardware triggered and Jetson Xavier AGX was switched to power mode with camera ROS2 driver and each module running respectively. \par For synchronization monitoring, the ROS2 message timestamps can be checked along with the ROS2 logger information and callbacks. For module running checking, callback functions with inference time and frames per second (FPS) could demonstrate the system performance. To adapt to the day and night light conditions, the camera parameters such as exposure time, gain, frame rate, etc. were tuned from the configuration files in order to capture the necessary scene information.

% \begin{figure}[ht]
%     \centering
%     \begin{subfigure}{.3\linewidth}
%         \centering
%         \includegraphics[width=\linewidth]{figs/resize_right_458.jpg}
%     \end{subfigure}
%     \begin{subfigure}{.3\linewidth}
%         \centering
%         \includegraphics[width=\linewidth]{figs/resize_right_459.jpg}
%     \end{subfigure}
%     \begin{subfigure}{.3\linewidth}
%         \centering
%         \includegraphics[width=\linewidth]{figs/resize_right_460.jpg}
%     \end{subfigure}

%     \hspace{0.5mm}

%     \centering
%     \begin{subfigure}{.3\linewidth}
%         \centering
%         \includegraphics[width=\linewidth]{figs/resize_left_433.jpg}
%     \end{subfigure}
%     \begin{subfigure}{.3\linewidth}
%         \centering
%         \includegraphics[width=\linewidth]{figs/resize_left_434.jpg}
%     \end{subfigure}
%     \begin{subfigure}{.3\linewidth}
%         \centering
%         \includegraphics[width=\linewidth]{figs/resize_left_435.jpg}
%     \end{subfigure}    

%     \caption{Captured sequential images of the recorded dataset: night driving scene (above), day driving scene (below).}
%     \label{fig:image_sequence}
% \vspace{-1em}
% \end{figure}

\subsection{Enhancement Results Comparison}

The experiment was conducted by estimating depth prediction and segmentation scores before and after the low light enhancement applied to nighttime dataset scenes. \par
\begin{figure}
\vspace{1em}
\centering
\includegraphics[width=0.4\textwidth]{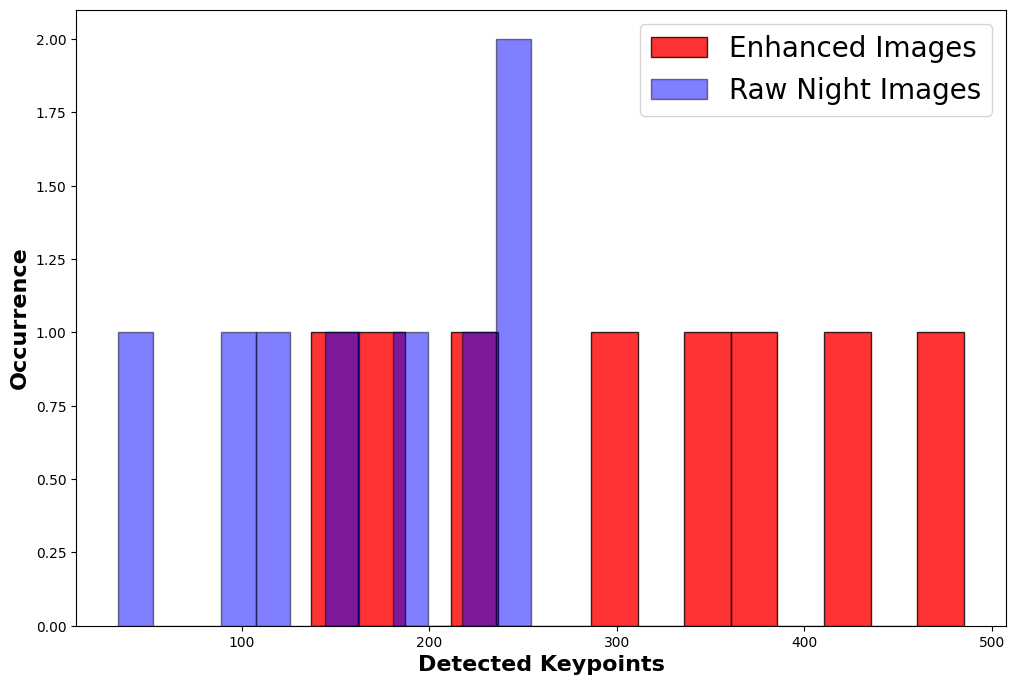}
\caption{Keypoints detection based on SIFT from night images and enhanced images.}
\label{fig:sift}
\vspace{0em}
\end{figure}

 First, the performance was evaluated on our recorded dataset. Scale Invariant Feature Transform (SIFT) presented by D. Lowe \cite{sift} has significant applications in object recognition and robot navigation. Feature analysis of 50 images from different critical driving scenes containing pedestrians, vehicles and traffic signs, etc. iconically was performed based on SIFT algorithms. The results in Fig. \ref{fig:sift} showed that on average, 148.1 more key points of objects in the images can be detected after enhancement. \par
\begin{figure}
    \centering
    \begin{subfigure}{.3\linewidth}
        \centering
        \includegraphics[width=\linewidth]{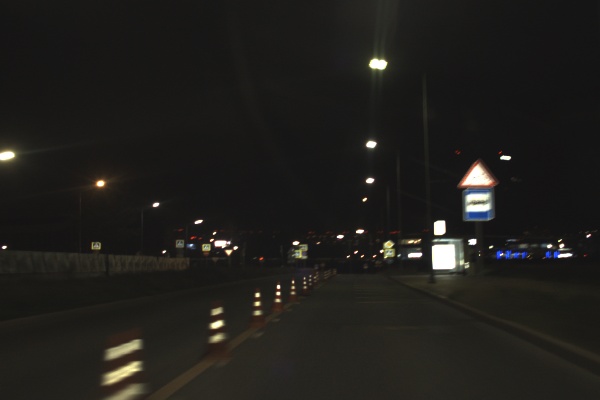}
    \end{subfigure}
    \begin{subfigure}{.3\linewidth}
        \centering
        \includegraphics[width=\linewidth]{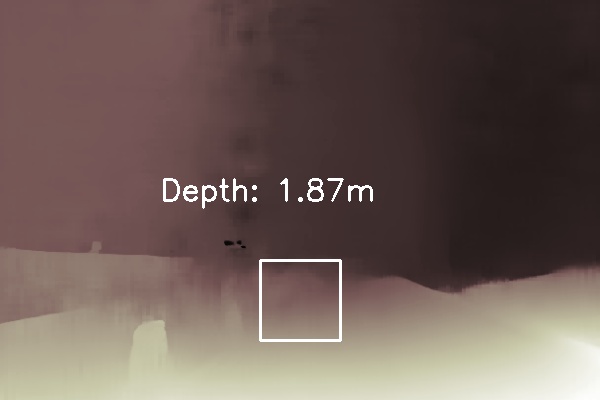}
    \end{subfigure}
    \begin{subfigure}{.3\linewidth}
        \centering
        \includegraphics[width=\linewidth]{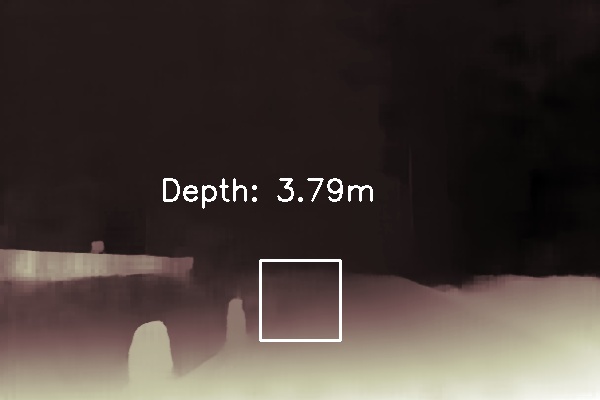}
    \end{subfigure}

    \hspace{0.5mm}

    \centering
    \begin{subfigure}{.3\linewidth}
        \centering
        \includegraphics[width=\linewidth]{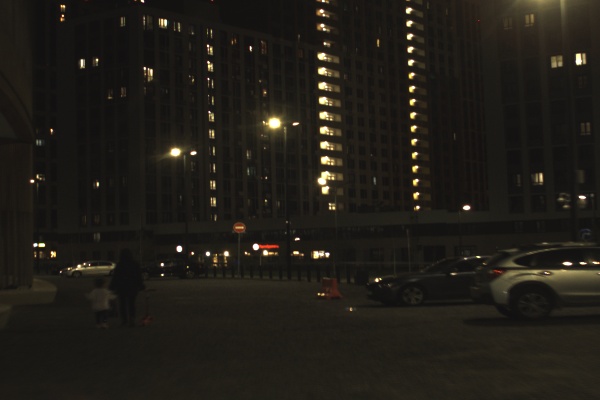}
        \subcaption{Input night \\ image.}
    \end{subfigure}
    \begin{subfigure}{.3\linewidth}
        \centering
        \includegraphics[width=\linewidth]{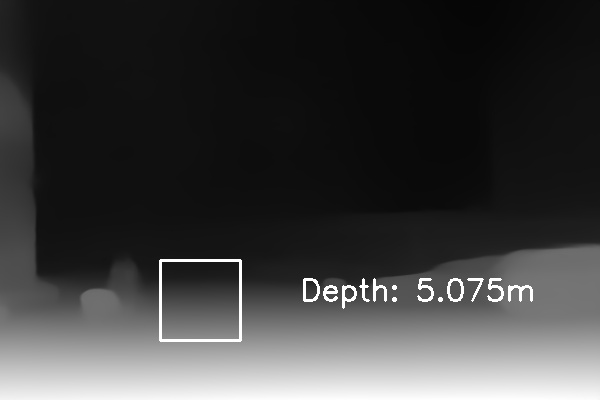}
        \subcaption{Output depth of the night image.}
    \end{subfigure}
    \begin{subfigure}{.3\linewidth}
        \centering
        \includegraphics[width=\linewidth]{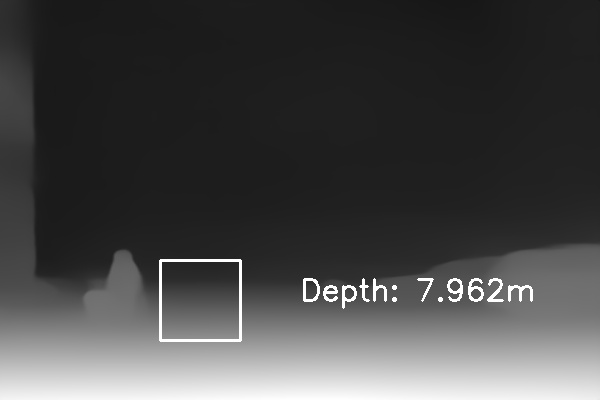}
        \subcaption{Output depth of the enhanced image.}
    \end{subfigure}
    
    \caption{Visual comparison of depth performance from Unimatch (above) and DPT (below) modules.}
    \label{fig:depth_visual_comparison}
\vspace{-1em}
\end{figure}

 For depth estimation performance, visual comparison between the night scene and the enhanced scene is provided in Fig. \ref{fig:depth_visual_comparison}, which demonstrates that due to the poor illumination, the detected targets did not display the necessary contour for recognition after the processing. However, the processed enhanced image improved the contour intactness. On our experimental self-driving car, Velodyne Ultra Puck (VLP-32) LiDAR is used to provide the ground truth of distance measurements. Obtaining the ground truth depth from annotated LiDAR point clouds shown in Fig. \ref{fig:LiDAR}, the errors between LiDAR and stereo camera depth estimation were reduced $27.16\%$ after the enhancement.

\begin{figure}
%\vspace{0em}
\centering
\includegraphics[width=0.47\textwidth]{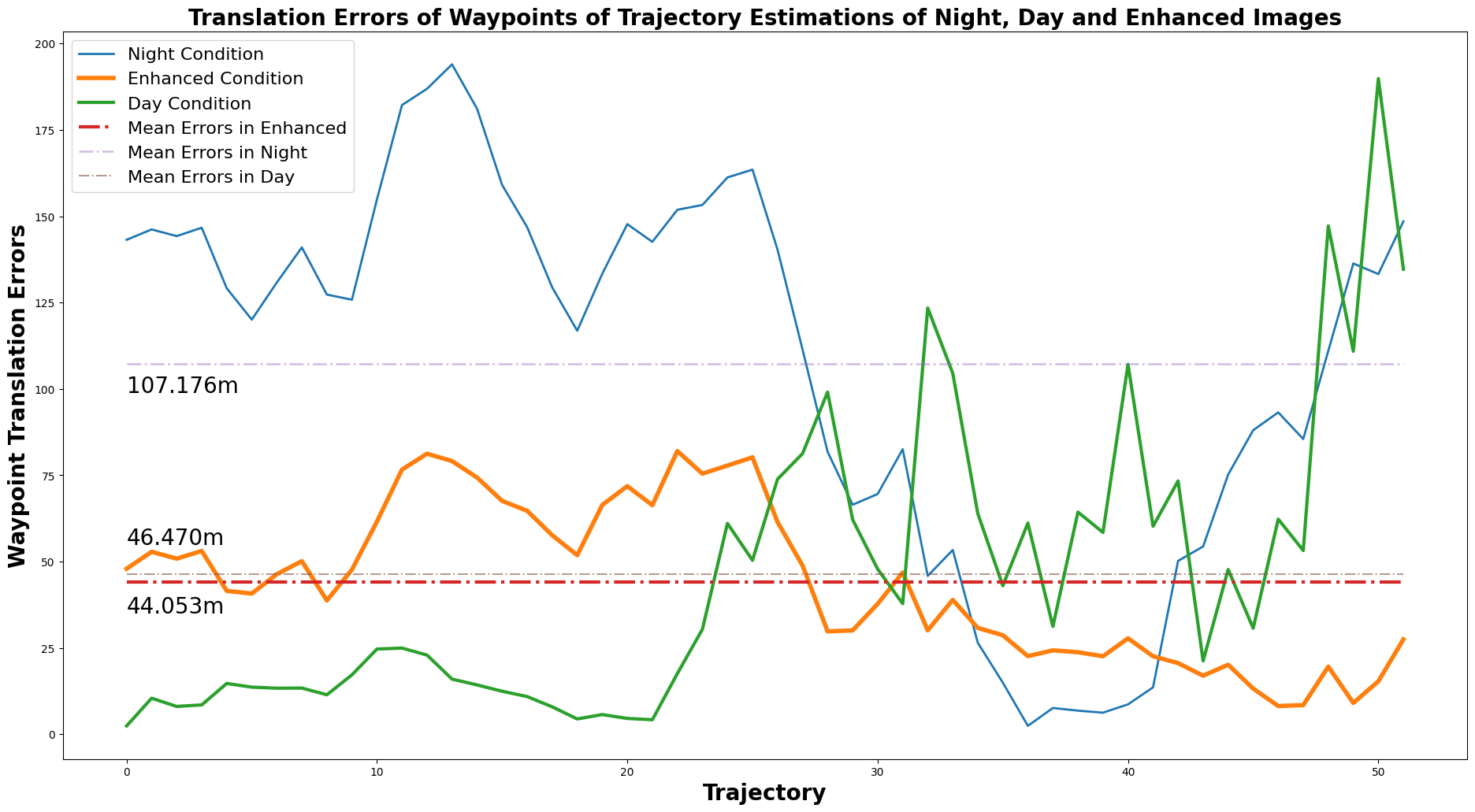}
\caption{Waypoints' translation errors of trajectory estimation on KITTI Odometry sequence 00 under day, night and enhanced conditions. The mean errors of night scene was $107.176$ m, while the mean errors of enhanced scene from our enhancement method reached $44.053$ m compared to $46.470$ m in day condition.}
\label{fig:vo}
\vspace{-1em}
\end{figure}

\vspace{-0.5em}
\begin{table}[h]
\centering
\caption{Average pixel accuracy comparison of our recorded dataset samples. Among 50 samples of critical scenes with intensive noise input, here is the showcase of 6 random samples out of 50 samples.}
\label{tab:Pix_acc_SegFormer}
\begin{tabular}{lp{2cm}|p{2cm}}
\multicolumn{1}{c}{}               & \multicolumn{1}{c|}{Night image} & \multicolumn{1}{c}{Enhanced image} \\ \hline
\multicolumn{1}{c}{Average pixel accuracy} & \multicolumn{1}{c|}{25.60\%}     & \multicolumn{1}{c}{\textbf{26.36\%}}  \\ \cline{2-3} 
\multicolumn{1}{c}{from sampling} & \multicolumn{1}{c|}{26.83\%}     & \multicolumn{1}{c}{\textbf{28.35\%}} \\ \cline{2-3} 
                                            & \multicolumn{1}{c|}{29.22\%}     & \multicolumn{1}{c}{\textbf{30.82\%}} \\ \cline{2-3} 
                                            & \multicolumn{1}{c|}{23.60\%}     & \multicolumn{1}{c}{\textbf{24.46\%}} \\ \cline{2-3}
                                            & \multicolumn{1}{c|}{26.67\%}     & \multicolumn{1}{c}{\textbf{30.90\%}} \\ \cline{2-3}
                                            & \multicolumn{1}{c|}{27.93\%}     & \multicolumn{1}{c}{\textbf{33.58\%}} 
\end{tabular}
\vspace{-1em}
\end{table}

\par In terms of semantic segmentation performance on our own dataset, the pixel accuracy \cite{moll2008truthing}, which denotes the percent of pixels that are accurately classified in the image based on the ground truth labels, was compared between corresponding night and enhanced images. Pixel accuracy, averaging over 50 processed images from different critical driving scenes with intensive noise input is provided in Table \ref{tab:Pix_acc_SegFormer}. It shows that enhanced images were $0.76\%$ higher on average pixel accuracy in comparison with dark ones. Visual comparison in Fig. \ref{fig:seg_visual_comparison} shows that a dark environment had a low level of contrast among objects within the scene in comparison with the enhanced one, which resulted in noisy pixel classification.

\subsection{Stereo Visual Odometry}

In driving scenes, long term performance over time based on image input such as visual odometry of large loop-closing should be evaluated. The experiment was conducted for estimating the driving trajectory under night, day and enhanced conditions based on feature tracking and pose estimation from stereo images. In Fig. \ref{fig:vo}, the waypoints along the ground truth of KITTI Odometry sequence 00 were selected for computing the translation errors between the waypoint coordinates of ground truth and estimation \cite{KITTI_Geiger2012CVPR}. Night scene of KITTI Odometry dataset was translated by an unsupervised image-to-image translation network while the enhanced scene was created based on our enhancement module from the synthetic night scene \cite{liu2017unsupervised}. \par The results showed that under night condition, the mean translation errors reached $107.176$ m along the whole sequence due to insufficient detected features and incontinuous tracking. After our enhancement, the mean translation errors were $2.417$ m less than ones in day condition.

\section{Conclusion}

A stereo visual perception system for a self-driving car, HawkDrive, with a processing pipeline running in the ROS2 framework has been developed. The pipeline can perform the hardware level synchronized image capturing from stereo cameras. Besides, throughout the developed modules, our stereo vision system can provide basic visual perception information such as depth estimation from the stereo camera and monocular camera, as well as semantic segmentation. To cope with poor light conditions during the driving scenes, a low light enhancement module has been developed to maintain nighttime driving safety and reliability. The experiments of the above development were conducted to verify the usage of all the modules. Thanks to richer detected features, in depth estimation tasks of stereo camera, our enhancement module can reduce the errors between LiDAR and stereo camera estimation by $27.16\%$. In semantic segmentation tasks, our enhancement was able to boost the pixel accuracy by $0.76\%$. Stereo visual odometry experiment verified the improvement on long term tasks by outperforming the trajectory estimation of day condition. \par There are still many topics that can be explored and researched. Utilizing the output information from our system remains an inspiring work, which could be integrated with LiDAR point cloud data to obtain a more precise and robust 3D perception scheme. Towards the hardware part, more Jetson devices and cameras can also be applied correspondingly to build a more practical and reliable perception scheme for self-driving cars.

\section*{Acknowledgment}

Research reported in this publication was financially supported by the Russian Science Foundation grant No. 24-41-02039 and the grant with India on the Swarm of Robots for Logistics. Our own dataset is captured in Skolkovo, Russia. The experimental car and devices are supported by Skolkovo Institute of Science and Technology, Moscow, Russia, and the LLC IntegraNT, Moscow, Russia. 

% The preferred spelling of the word ÒacknowledgmentÓ in America is without an ÒeÓ after the ÒgÓ. Avoid the stilted expression, ÒOne of us (R. B. G.) thanks . . .Ó  Instead, try ÒR. B. G. thanksÓ. Put sponsor acknowledgments in the unnumbered footnote on the first page.

% %%%%%%%%%%%%%%%%%%%%%%%%%%%%%%%%%%%%%%%%%%%%%%%%%%%%%%%%%%%%%%%%%%%%%%%%%%%%%%%%

% References are important to the reader; therefore, each citation must be complete and correct. If at all possible, references should be commonly available publications.

\bibliographystyle{IEEEtran}
\bibliography{references}

\end{document}